\documentclass{ifacconf}

\usepackage{graphicx}      
\usepackage{natbib}        

\usepackage{amsmath}

\begin{document}
\begin{frontmatter}

\title{Dynamic Modeling and Robust Gait Optimization of a Compliant Worm Robot\thanksref{footnoteinfo}} 


\thanks[footnoteinfo]{This work was supported by National Science Foundation (CNS 2125484).}

\author[First]{Xinyu Zhou}, 
\author[First]{Yu Mei}, 
\author[First]{Faith Thomson}, 
\author[First]{Christian Luedtke}, 
\author[First]{Xinda Qi}, and
\author[First]{Xiaobo Tan}

\address[First]{Michigan State University, 
   East Lansing, MI 48824 USA\\ (e-mail: {zhouxi63, meiyu1, thomsonf, luedtke2, qixinda, xbtan}@msu.edu)}

\begin{abstract}                
Worm-inspired robots provide an effective locomotion strategy for constrained environments by combining cyclic body deformation with alternating anchoring. For compliant robots, however, the interaction between deformable anchoring structures and the environment makes predictive modeling and deployable gait optimization challenging. This paper presents an experimentally grounded modeling and optimization framework for a compliant worm robot capable of traversing corrugated pipes. First, a hybrid dynamic locomotion model is derived, in which the robot motion is represented by continuous dynamics within a corrugation groove and discrete switching of anchoring positions between adjacent grooves. A slack-aware actuation model is further introduced to map the commanded gait input to the realized body-length change, and an energy model is developed based on physics and calibrated with empirical power measurement. Based on these models, a multi-objective gait optimization problem is formulated to maximize average speed while minimizing average power. To reduce the fragility of nominal boundary-seeking solutions, a kinematic robustness margin is introduced into the anchoring-transition conditions, leading to a margin-based robust gait optimization framework. Experimental results show that the proposed framework captures the dominant locomotion and energy-consumption behavior of the robot over the tested conditions, and enables robust gait optimization for achieving speed-power trade-off.
\end{abstract}

\begin{keyword}
Compliant robot; constrained-environment locomotion; worm robot; modeling; gait optimization
\end{keyword}

\end{frontmatter}


\section{Introduction}
\label{sec:introduction}
Worm-inspired robots provide an effective locomotion strategy for confined and cluttered environments, where cyclic body deformation and alternating anchoring can generate net motion with relatively simple mechanical structures. Such robots have been explored for applications including locomotion on complex terrain \citep{das2023terrain, gu2024terrain, horchler2015terrain, moreira2018terrain, niu2021terrain, onal2012origamiTiltedLeg, joey2019inchwormFeature}, traversal of tubular environments \citep{tao2024tubular, li2019tubular, zhang2019tubular, liu2022tubular, fang2023tubular}, and wall climbing \citep{yang2018climbing, zhang2021climbing, zheng2018climbing, wang2008climbing}. Many worm-inspired robots achieve directional locomotion through anisotropic interaction with the environment. In compliant designs, this anisotropy is often realized through deformable anchoring structures whose resistance depends on the direction of motion \citep{onal2012origamiTiltedLeg, luedtke20253d}. Such designs are attractive because the compliant interaction can better accommodate geometric variation and environmental complexity. However, the same compliance that enables locomotion also makes prediction more difficult: deformation of the anchoring structure, finite clearance, and anchoring transition sensitivity can all introduce discrepancies between commanded actuation and realized locomotion response. These effects become especially important when the predictive modeling is intended to support gait optimization and hardware experiments.
To support systematic gait design, appropriate mathematical models are needed. Kinematic models under ideal anchoring assumptions are useful for preliminary analysis and design intuition \citep{onal2012origamiTiltedLeg, luedtke20253d}, but they are generally insufficient when locomotion depends strongly on compliant environmental interaction and nonideal anchoring transitions. In such cases, the robot motion typically involves both continuous dynamics and discrete switching events, which motivates dynamic or hybrid-system representations. More importantly, for hardware-valid gait optimization, locomotion modeling alone is not enough: one must also account for how the commanded actuation is transmitted to the robot body and how the resulting motion translates into energetic cost.

In this paper, we investigate an experimentally grounded modeling and optimization framework with a compliant worm robot designed for traversing corrugated pipes. We first derive hybrid dynamic locomotion model, in which the robot motion is represented by continuous dynamics within a corrugation groove and discrete switching of anchoring positions between adjacent grooves. The locomotion model propose a clearance-aware fin--groove anisotropic interaction law, which helps capture the anisotropic motion principle and dynamic motion of the worm robot. We used the robot's body length change as the input of the locomotion model. We then introduce a slack-aware first-order actuation model to map the commanded gait input to the realized body-length change, and formulate an energy model from physics and calibrated with empirical power measurement. Finally, we integrate these models into a robust multi-objective gait optimization framework with a kinematic robustness margin. Experimental results on verifying the modeling response and deploying optimized gait validate the proposed modeling can capture the dominant locomotion and energy-consumption behavior of the robot, and show that the robust optimizing framework enables deployable gait optimization for achieving speed--power trade-off.

    
    

The design and fabrication of the robot platform for corrugated pipes was initially introduced in \citep{luedtke20253d}. A preliminary version of part of this work was presented at the American Control Conference (ACC) \citep{zhou2025dynamic}, where a nominal hybrid dynamic locomotion model was proposed and used for initial simulation-based analysis. Compared with that ACC paper, the present work adds experimental identification and validation of the locomotion model, a slack-aware actuation model that accounts for the discrepancy between commanded and realized body-length change, an energy model that evaluates the energy consumption cost during locomotion, and a robust gait optimization framework with hardware validation.

The remainder of this paper is organized as follows. Section~II introduces the robot system and the overall problem formulation. Section~III presents the proposed modeling of the worm robot, including the locomotion, actuation, and energy models. Section~IV develops the optimization approach with a kinematic robustness margin. Section~V reports the experimental parameter identification and validation results for the models, together with the hardware validation of optimized gaits. Finally, Section~VI concludes the paper.


\section{Robot System and Problem Formulation}
\label{sec:robot_system_problem}

\begin{figure}[t]
    \centering
     \includegraphics[width=0.82\columnwidth]{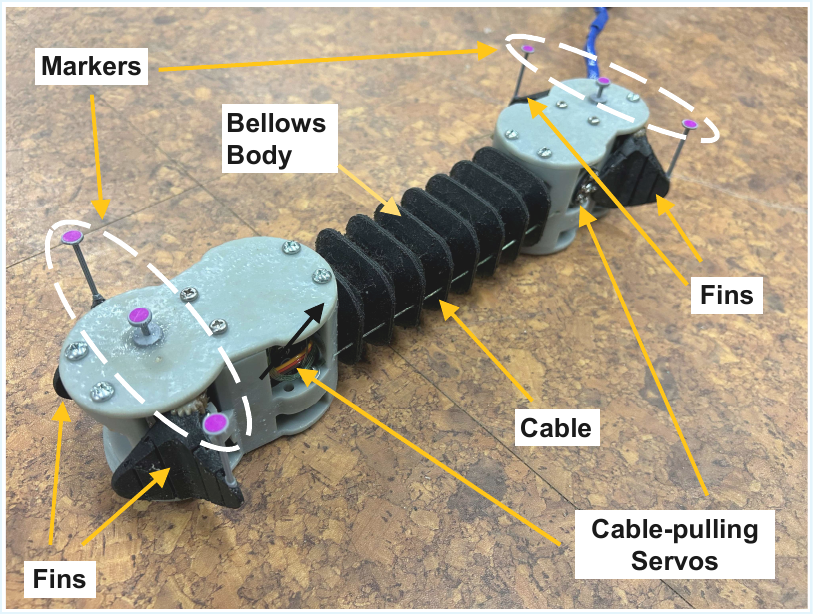}
    \caption{Prototype of the compliant worm robot. The robot consists of a compliant bellows body and two end fin modules. The visual markers are attached for robot motion tracking and are not essential to the robot's locomotion mechanism.}
    \label{fig:robot_prototype}
\end{figure}

\begin{figure}[t]
    \centering
    \includegraphics[width=0.80\columnwidth]{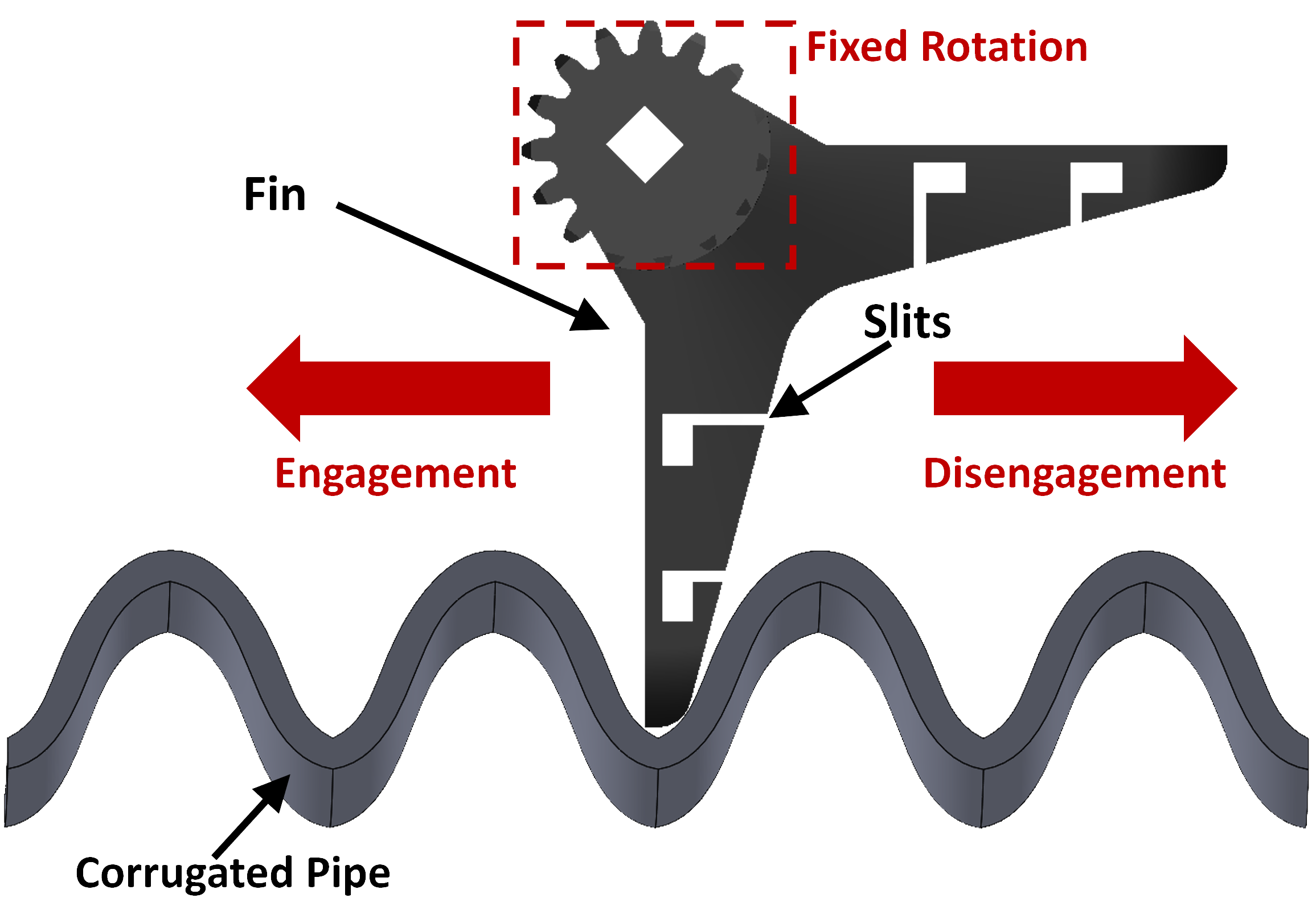}
    \caption{Illustration of the structure of a single fin and its interaction with the corrugated pipe. Each fin comprises an embedded gear driven by the servo inside the fin module and two compliant bars used for interaction with the corrugated pipe. During operation, one compliant bar interacts with the pipe wall and achieves anchoring engagement or anchoring disengagement depending on the locomotion direction.}
    \label{fig:fin_intro}
\end{figure}

\subsection{Problem Formulation}
\label{subsec:problem_formulation}

\begin{figure*}[t]
    \centering
    \includegraphics[width=0.9\textwidth]{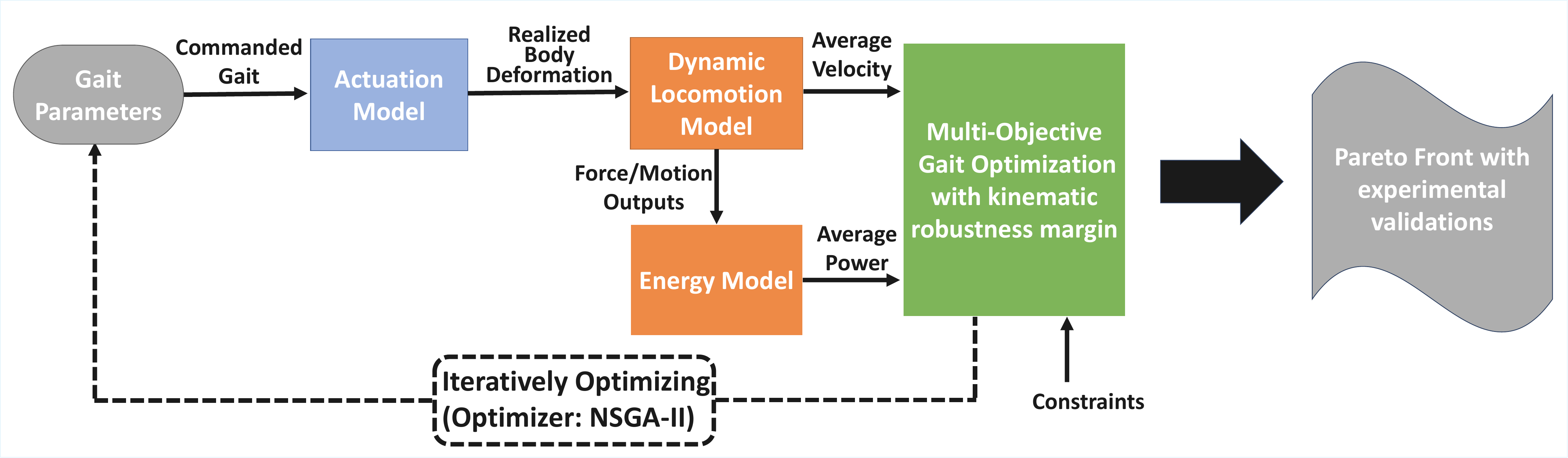}
    \caption{System-level view of the problem formulation considered in this paper. A commanded gait, parameterized by stroke and frequency, is mapped by the actuation model to the realized body-length trajectory, which then drives the locomotion and energy models used for optimization.}
    \label{fig:problem_formulation_pipeline}
    \vspace{-2mm}
\end{figure*}

\subsection{Introduction to the Robot System}
\label{subsec:robot_system}

The compliant worm robot studied in this work follows the general design principle developed in a previous work \citep{luedtke20253d}. A prototype of the robot is shown in Fig.~\ref{fig:robot_prototype}. The robot consists of a compliant central body and two fin modules located at its ends. The central body is formed by a bellows structure, which provides the axial compliance required for cyclic contraction and expansion. This body deformation is driven by a cable transmission actuated by servo motors housed in the end modules. As the robot body contracts and expands, the robot interacts with the corrugated pipe through the compliant and anisotropic fins and generates directional locomotion. 

Fig.~\ref{fig:fin_intro} illustrates the structure of a single fin and its interaction with the corrugated pipe. Each fin consists of an embedded gear driven by the servo inside the fin module and two compliant bars used for interaction with the pipe wall. During operation, only one compliant bar interacts with the pipe wall, while the other remains clear, as shown in the figure. One side of the compliant bar features slits, resulting in anisotropic behavior. When the fin moves toward the side without slits, it becomes difficult to deform sufficiently to navigate past the pipe's ridges, allowing it to anchor at its current position, referred to as ``anchoring engagement''. Conversely, when the fin moves toward the side with slits, the deformation stiffness decreases, enabling it to easily pass the ridges and transition to the next position, termed ``anchoring disengagement''. As the robot's body contracts and expands, the fin alternates between anchoring engagement and disengagement, facilitating locomotion in one direction. Additionally, the fin modules can flip the fins via their servos, allowing the other compliant bar to interact with the pipe, thereby reversing the anisotropic properties and the locomotion direction. In this paper, we focus on the forward-locomotion case, since the reverse-locomotion case differs mainly by direction reversal.

For locomotion experiments, visual markers are attached to the robot body and end modules to facilitate motion tracking. These markers are used only for experimental measurement and are not essential to the robot’s locomotion mechanism.

The objective of this work is to develop an experimentally grounded modeling and optimization framework for the compliant worm robot in corrugated pipes. Given a periodic commanded gait, the robot undergoes cyclic body deformation, interacts with the corrugated pipe through anisotropic fin anchoring, and produces net locomotion. The central challenge is that predictive gait optimization must account not only for the hybrid locomotion mechanics, but also for the discrepancy between commanded actuation and realized body-length change, together with the associated energy consumption at the system level. Therefore, the goal is to establish a framework that connects commanded actuation, realized body-length change, locomotion response, and energy cost in a form suitable for optimization and hardware validation.

Let $L(t)=L_0+\Delta L(t)$ denote the realized robot body length, defined as the distance between the roots of the two fin pairs, where $L_0$ is the initial robot body length and $\Delta L(t)$ is the realized body-length change. Under the sign convention adopted in this work, contraction corresponds to negative body-length change, i.e., $\Delta L(t)\le 0$ during contraction-dominant operation. The commanded gait is parameterized by the contraction stroke $S$ and the operation frequency $f$. In the optimization stage, the commanded actuation is taken as the sinusoidal input
\begin{equation}
    u_{\mathrm{cmd}}(t)
    =
    -\frac{S}{2}\left[1-\cos(2\pi f t)\right],
    \qquad
    \Delta L(0)=0,
    \label{eq:problem_formulation_gait}
\end{equation}
where the negative sign is consistent with the contraction-positive-to-negative convention used throughout the paper.

Fig.~\ref{fig:problem_formulation_pipeline} summarizes the problem formulation adopted in this paper. For a given gait parameter pair $(S,f)$, the commanded input $u_{\mathrm{cmd}}(t)$ is first mapped by the actuation model to the realized body-length change $\Delta L(t)$ and the realized body length $L(t)$. This realized body-length trajectory is then supplied to the hybrid locomotion model to predict the locomotion response in the corrugated pipe, and together with the locomotion states it is used in the energy model to predict the corresponding power consumption. In the optimization stage, each candidate gait is evaluated using the average locomotion speed $\bar{v}$ and the average power consumption $\bar{P}$.

Accordingly, the overall problem considered in this paper is to use the pipeline in Fig.~\ref{fig:problem_formulation_pipeline} to predict the locomotion performance and energy consumption associated with a candidate gait, and to identify gait parameters that improve the trade-off between average speed and average power in the practical robot system. To address this problem, Section~III formulates the theoretical models of the robot, including the hybrid locomotion model, a slack-aware actuation model, and the energy model. Section~IV develops the robust multi-objective optimization approach with a kinematic margin. Section~V then presents the experimental parameter identification and validation results for the models, together with the hardware validation of selected optimized gaits.


\section{Theoretical Models of the Worm Robot}
\label{sec:theoretical_models}

This section develops the three theoretical components used in the optimization framework. The hybrid locomotion model predicts robot motion from the realized body-length input, the actuation model maps the commanded input to that realized body-length change, and the energy model estimates system power consumption from the predicted motion and actuation response.

\subsection{Hybrid Dynamic Locomotion Model}
\label{subsec:baseline_hybrid_model}

\begin{figure}[t]
    \centering
    \includegraphics[width=0.9\linewidth]{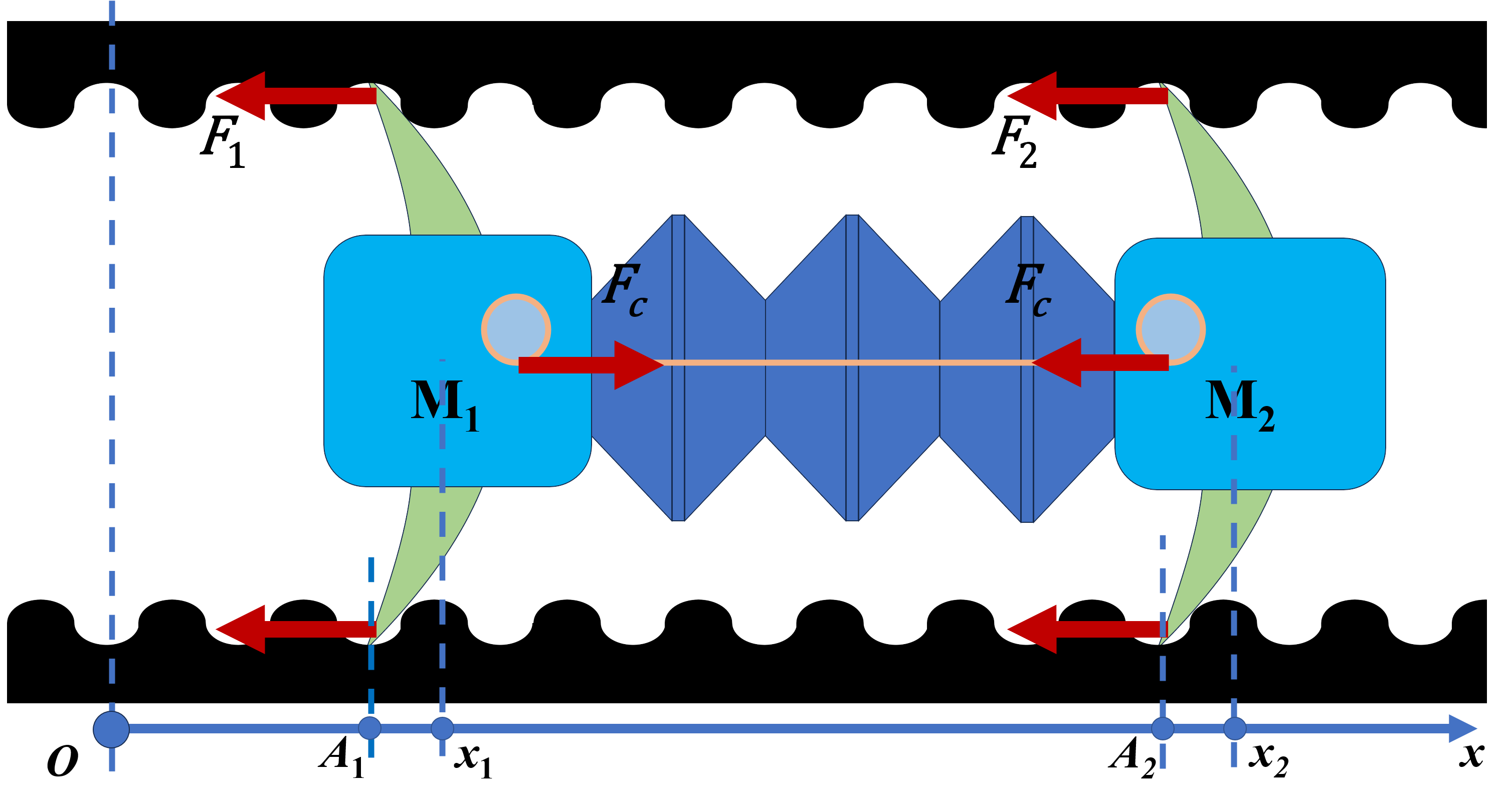}
    \caption{A schematic diagram of the worm robot moving in a corrugated pipe. $M_1$ and $M_2$ are the designation of the two mass blocks, as well as their masses. $F_1$, $F_2$, $F_c$ represent the forces acting on $M_1$ and $M_2$ through the fins, and the traction force of the cable, respectively. $x_1$ and $x_2$ are the positions of the two mass blocks, in more details, the positions of where the fins' roots are located, while $A_1$ and $A_2$ are the anchoring positions between each pair of fins and the corrugated pipe.}
    \label{fig:locomotion_schematic}
\end{figure}

The locomotion of the compliant worm robot in a corrugated pipe is represented in the form of a hybrid system, consisting of continuous motion within a groove and discrete switching of anchoring positions between adjacent grooves. A schematic of the robot platform is shown in Fig.~\ref{fig:locomotion_schematic}. The two end modules of the robot are represented by two lumped masses, denoted by $M_1$ and $M_2$, with positions $x_1(t)$ and $x_2(t)$, respectively. The corresponding anchoring positions of the two fin pairs in the corrugated pipe are denoted by $A_1$ and $A_2$. The compliant bellows body connecting the two end modules is modeled by an equivalent spring--damper pair with stiffness coefficient $k_{\mathrm{b}}$ and damping coefficient $c_{\mathrm{b}}$. In addition, the environmental dissipation is represented by a lumped viscous coefficient $\eta$.

Let $F_{\mathrm{c}}$ denote the cable traction force, and let $F_1$ and $F_2$ denote the interaction forces between the two fin pairs and the corrugated pipe. The original continuous dynamics of the two-mass system are written as
\begin{equation}
    M_1 \ddot{x}_1
    =
    F_{\mathrm{c}}
    +
    k_{\mathrm{b}}(x_2-x_1-L_{\mathrm{free}})
    +
    c_{\mathrm{b}}(\dot{x}_2-\dot{x}_1)
    -
    \eta \dot{x}_1
    -
    F_1,
    \label{eq:x1_ddot_full}
\end{equation}
\begin{equation}
    M_2 \ddot{x}_2
    =
    -F_{\mathrm{c}}
    -
    k_{\mathrm{b}}(x_2-x_1-L_{\mathrm{free}})
    -
    c_{\mathrm{b}}(\dot{x}_2-\dot{x}_1)
    -
    \eta \dot{x}_2
    -
    F_2,
    \label{eq:x2_ddot_full}
\end{equation}
where $L_{\mathrm{free}}$ is the unstressed robot body length.

The fin--pipe interaction forces are modeled as
\begin{equation}
    F_1 = f(x_1-A_1), \qquad
    F_2 = f(x_2-A_2),
    \label{eq:F1_F2}
\end{equation}
where $f(\cdot)$ is the fin--groove interaction function. In the present work, $f(\cdot)$ is modeled as a clearance-aware piecewise linear function:
\begin{equation}
    f(x)=
    \begin{cases}
        k_{\mathrm{eng}}\left(x-\dfrac{\delta_{\mathrm{c}}}{2}\right), & x>\dfrac{\delta_{\mathrm{c}}}{2},\\[6pt]
        0, & -\dfrac{\delta_{\mathrm{c}}}{2}\le x \le \dfrac{\delta_{\mathrm{c}}}{2},\\[6pt]
        k_{\mathrm{dis}}\left(x+\dfrac{\delta_{\mathrm{c}}}{2}\right), & x<-\dfrac{\delta_{\mathrm{c}}}{2},
    \end{cases}
    \label{eq:fin_force_piecewise}
\end{equation}
where $\delta_{\mathrm{c}}$ denotes the effective clearance width of the fin--groove interaction, and $k_{\mathrm{eng}}$ and $k_{\mathrm{dis}}$ are the effective directional stiffness coefficients associated with anchoring engagement and disengagement, respectively. The difference between $k_{\mathrm{eng}}$ and $k_{\mathrm{dis}}$ captures the anisotropy of the system.

To match the actuation-oriented formulation used in the later sections, the locomotion model is rewritten using the realized robot body length
\begin{equation}
    L(t)=L_0+\Delta L(t),
    \label{eq:L_definition_locomotion}
\end{equation}
as the system input, where $\Delta L(t)$ is the realized body-length change. Since
\begin{equation}
    x_2(t)-x_1(t)=L(t),
    \label{eq:length_constraint}
\end{equation}
the state dimension can be reduced from the original four-state representation to a two-state model. Let
\begin{equation}
    \mathbf{x}(t)=
    \begin{bmatrix}
        x_1(t)\\
        \dot{x}_1(t)
    \end{bmatrix}.
\end{equation}
Substituting \eqref{eq:length_constraint} into \eqref{eq:x1_ddot_full}--\eqref{eq:x2_ddot_full} and combining the two equations yields the reduced continuous dynamics

\begin{equation}
\begin{split}    
    \dot{\mathbf{x}} &= \mathbf{F}_{A_1,A_2} (\mathbf{x},L(t))\\
            & = \begin{bmatrix}
\dot{x}_1 \\\\
\frac{1}{M_1+M_2}[-2 \eta \dot{x}_1-f(x_1-A_1)-\\
f(x_1+L-A_2)-\eta \dot{L}-M_2 \ddot{L}] 
\end{bmatrix} 
\end{split},
    \label{eq:continuous_model}
\end{equation}

The hybrid nature of the model arises from the fact that the anchoring positions $A_1$ and $A_2$ remain fixed while the corresponding fins stay within the same groove, but update discretely once a switching condition is met. Let $d$ denote the corrugated-pipe pitch and let $p_{\mathrm{sw}}$ denote the switching threshold. The switching logic governing the anchoring-position updates are described by
\begin{equation}
    A_1(t^+)=
    \begin{cases}
        A_1(t^-)+d, & x_1(t^-)-A_1(t^-)>p_{\mathrm{sw}},\\
        A_1(t^-)-d, & x_1(t^-)-A_1(t^-)<-p_{\mathrm{sw}},\\
        A_1(t^-), & \text{otherwise},
    \end{cases}
    \label{eq:A1_switch}
\end{equation}
\begin{equation}
    A_2(t^+)=
    \begin{cases}
        A_2(t^-)+d, & x_1(t^-)+L(t^-)-A_2(t^-) > p_{\mathrm{sw}},\\
        A_2(t^-)-d, & x_1(t^-)+L(t^-)-A_2(t^-) < -p_{\mathrm{sw}},\\
        A_2(t^-), & \text{otherwise}.
    \end{cases}
    \label{eq:A2_switch}
\end{equation}
Equations~\eqref{eq:continuous_model}--\eqref{eq:A2_switch} together define the baseline hybrid locomotion model used in this work.

Once $x_1(t)$ is obtained from \eqref{eq:continuous_model}--\eqref{eq:A2_switch}, the remaining kinematic outputs follow from \eqref{eq:length_constraint}, namely, $x_2(t)=x_1(t)+L(t)$ and $\dot{x}_2(t)=\dot{x}_1(t)+\dot{L}(t)$. The cable force is then recovered as
\begin{equation}
    F_{\mathrm{c}}
    =
    M_1\ddot{x}_1
    +
    k_{\mathrm{b}}(L_{\mathrm{free}}-L(t))
    -
    c_{\mathrm{b}}\dot{L}(t)
    +
    \eta \dot{x}_1
    +
    f(x_1-A_1).
    \label{eq:Fc}
\end{equation}

It is worth noting that the bellows damping coefficient $c_{\mathrm{b}}$ appears in the original two-mass dynamics and in the cable-force expression, but does not explicitly enter the reduced locomotion model once the system is reformulated with the realized body length $L(t)$ as input. Therefore, $c_{\mathrm{b}}$ is not identified in the locomotion-model identification stage and is instead identified later in the energy-model identification.


\subsection{Slack-Aware Actuation Model}
\label{sec:actuation_modeling}

The hybrid locomotion model is driven by the realized robot body length in \eqref{eq:L_definition_locomotion}, whereas the controller specifies the commanded length-change input $u_{\mathrm{cmd}}(t)$. In hardware, servo response lag, cable transmission slack and structural compliance make the realized response differ from the commanded one, especially near motion reversal. Since this actuation layer is introduced only to bridge the controller command and the realized body-length change, a compact reduced-order description is adopted here.

In the periodic operating regime considered in this work, the reversal-dependent slack is represented by a one-sided slack clip,
\begin{equation}
u_{\mathrm{eff}}(t)=\min\!\left(u_{\mathrm{cmd}}(t),-\delta_{\mathrm{s}}\right),
\label{eq:slack_clip}
\end{equation}
where $u_{\mathrm{eff}}(t)$ is the effective input after slack clipping and $\delta_{\mathrm{s}}>0$ is the identified slack width. Under the contraction-negative sign convention, both $u_{\mathrm{cmd}}(t)$ and $u_{\mathrm{eff}}(t)$ are non-positive. Equation~\eqref{eq:slack_clip} means that, after reversal, the effective input remains clipped at the preload level $-\delta_{\mathrm{s}}$ until the commanded contraction exceeds the slack width again; away from this clipped interval, $u_{\mathrm{eff}}(t)$ follows $u_{\mathrm{cmd}}(t)$.

The realized body-length change is then modeled by a first-order linear dynamics:
\begin{equation}
\tau \dot{\Delta L}(t) + \Delta L(t) = K\,u_{\mathrm{eff}}(t),
\label{eq:first_order_actuation}
\end{equation}
where $\tau>0$ is the actuation time constant and $K>0$ is the steady-state transmission gain. Together, \eqref{eq:slack_clip} and \eqref{eq:first_order_actuation} map $u_{\mathrm{cmd}}(t)$ to $\Delta L(t)$, while the realized robot body length supplied to the locomotion model follows from \eqref{eq:L_definition_locomotion}. The parameters to be identified in the actuation model are $\boldsymbol{\theta}_{\mathrm{act}}=[\delta_{\mathrm{s}},\;K,\;\tau]^\top$.


\subsection{Energy Model}
\label{sec:energy_modeling}
\label{subsec:energy_model_formulation}

To complete the gait optimization framework, the energy model is further developed based on physics and calibrated with empirical power measurement. The formulation starts from the positive mechanical cable power predicted by the locomotion model and introduces a lumped actuation power factor to account for the difference between idealized mechanical power and the source-side power measured from the physical robot, thereby capturing actuator inefficiency, transmission loss, and other unmodeled energy loss in a compact form.

The energy consumption of the worm robot over time is related to the positive mechanical power delivered by the cable-pulling servo motors \citep{xi2016optimization}. Motivated by this observation, the total system energy consumption power is modeled as
\begin{equation}
    P(t) = P_{\mathrm{idle}} + \alpha_{\mathrm{P}} \cdot \max\!\left(-F_{\mathrm{c}}(t)\dot{L}(t),\,0\right),
    \label{eq:power_model}
\end{equation}
where $P_{\mathrm{idle}}$ is the baseline idle power of the robot system, $\alpha_{\mathrm{P}}$ is a lumped actuation power factor, $F_{\mathrm{c}}(t)$ is the cable force, and $\dot{L}(t)$ is the robot body-length rate. From \eqref{eq:L_definition_locomotion}, one has $\dot{L}(t)=\dot{\Delta L}(t)$. Under the sign convention adopted in this work, contraction corresponds to negative body-length change, so that the term $-F_{\mathrm{c}}(t)\dot{L}(t)$ represents the positive mechanical cable power during contraction-dominant operation.

The model in \eqref{eq:power_model} consists of a baseline term and an actuation-dependent term. The baseline term $P_{\mathrm{idle}}$ accounts for the electrical power required to maintain system operation even when no effective positive cable work is performed. The actuation-dependent term scales the positive mechanical cable power by the lumped factor $\alpha_{\mathrm{P}}$, which bridges the idealized mechanical work term and the source-side electrical power measured in the experiments. As a result, $\alpha_{\mathrm{P}}$ absorbs the combined effects of actuation inefficiency, transmission loss, and other unmodeled energy dissipation in the physical system.




\section{Optimization Approach}
\label{sec:optimization_approach}

With the theoretical models developed in Section~III, the remaining task is to optimize the gait parameters for improved locomotion performance and energetic efficiency. A direct application of nominal model-based optimization, however, tends to yield solutions that are too close to the anchoring-transition boundary and therefore fragile in hardware experiments. To mitigate this issue, a kinematic robustness margin is introduced into the anchoring-switching thresholds, leading to a margin-based robust gait optimization framework. In this section, only the optimization method is described. The numerical selection of the robustness margin, which is carried out after model identification, is reported later in Section~V.

\subsection{Kinematic Robustification}
\label{subsec:robustness_margin_selection}

Let $\delta_m$ denote the imposed kinematic robustness margin. In the baseline locomotion model, the anchoring-position updates are governed by the switching threshold $p_{\mathrm{sw}}$ in \eqref{eq:A1_switch}--\eqref{eq:A2_switch}. In the robust formulation, this threshold is shifted conservatively by $\delta_m$, so that the switching conditions become
\begin{equation}
    A_1(t^+)=
    \begin{cases}
        A_1(t^-)+d, & x_1(t^-)-A_1(t^-) > p_{\mathrm{sw}}+\delta_m,\\
        A_1(t^-)-d, & x_1(t^-)-A_1(t^-) < -(p_{\mathrm{sw}}+\delta_m),\\
        A_1(t^-), & \text{otherwise},
    \end{cases}
    \label{eq:A1_switch_robust}
\end{equation}
\begin{equation}
    A_2(t^+)=
    \begin{cases}
        A_2(t^-)+d, & x_2(t^-)-A_2(t^-) > p_{\mathrm{sw}}+\delta_m,\\
        A_2(t^-)-d, & x_2(t^-)-A_2(t^-) < -(p_{\mathrm{sw}}+\delta_m),\\
        A_2(t^-), & \text{otherwise}.
    \end{cases}
    \label{eq:A2_switch_robust}
\end{equation}
Equivalently, the robust formulation can be interpreted as replacing the nominal switching threshold $p_{\mathrm{sw}}$ by an enlarged threshold $p_{\mathrm{sw}}+\delta_m$. As a result, the optimizer is prevented from selecting gaits that only marginally satisfy the nominal transition condition.

The value of $\delta_m$ is not prescribed a priori. Instead, after the locomotion, actuation, and energy models have been identified, it is selected empirically through a price-of-robustness analysis. Specifically, a range of candidate margin values is imposed, and for each candidate value the corresponding optimal cost of transport (COT) is computed. Although the final optimization objectives in this work are average speed and average power, the optimal COT provides a convenient scalar indicator for identifying when the imposed conservatism begins to introduce a significant performance penalty. The selected margin is then taken at the cliff point of the resulting price-of-robustness curve. The corresponding analysis and the selected value used in the final experiments are reported in Section~V; see Fig.~\ref{fig:robustness_margin_scan}.

\subsection{Multi-Objective Gait Optimization}
\label{subsec:optimization_formulation}

For a fixed choice of $\delta_m$, the gait optimization is formulated as a multi-objective problem. The decision vector is $\mathbf{p} = [S,\;f]^\top$, where $S$ is the contraction stroke and $f$ is the operation frequency.

For each candidate gait $\mathbf{p}$, the commanded actuation follows the sinusoidal input in \eqref{eq:problem_formulation_gait}. The commanded input is first mapped to the effective input and realized body-length trajectory by the slack-aware actuation model. The resulting body-length trajectory is then supplied to the hybrid locomotion model, where the switching conditions are modified according to \eqref{eq:A1_switch_robust}--\eqref{eq:A2_switch_robust}. Based on the predicted locomotion and energy responses, the average speed and average power are evaluated.

The two optimization objectives considered in this work are the average locomotion speed and the average power consumption. For each candidate gait, the robot motion is simulated over five actuation cycles, and the last three cycles are used to compute the performance metrics so as to reduce the influence of initial transients. Let $\bar{v}(\mathbf{p})$ denote the average speed computed from the last three simulated cycles, and let $\bar{P}(\mathbf{p})$ denote the corresponding average power over the same interval.

The multi-objective optimization problem is then formulated as
\begin{equation}
    \min_{\mathbf{p}}
    \;\;
    \Big[
    -\bar{v}(\mathbf{p}),
    \;
    \bar{P}(\mathbf{p})
    \Big],
    \label{eq:moo_formulation}
\end{equation}
subject to
\begin{equation}
    \mathbf{p}_{\min} \le \mathbf{p} \le \mathbf{p}_{\max},
\end{equation}
together with the robust switching conditions in \eqref{eq:A1_switch_robust}--\eqref{eq:A2_switch_robust}. The optimization problem in \eqref{eq:moo_formulation} is solved using NSGA-II, yielding a Pareto set of optimized sinusoidal gaits.



\section{Experimental Parameter Identification and Results}
\label{sec:experimental_validation}

This section follows the logic of Sections~III and IV. It reports experimental parameter identification and representative results for the locomotion, actuation, and energy models, then presents the price-of-robustness analysis used to select the kinematic robustness margin, and finally reports hardware results for representative optimized gaits.

\begin{table}[htbp]
\caption{Parameters of the Worm Robot System}
\begin{center}
\begin{tabular}{lcc}
\hline
Parameter & Symbol & Value \\
\hline
\multicolumn{3}{c}{\textbf{Robot Body Dynamics}} \\
\hline
Mass of the rear body segment & $M_1$ & 0.429 kg \\
Mass of the front body segment & $M_2$ & 0.429 kg \\
Unstressed length of robot body & $L_{\mathrm{free}}$ & 0.30 m \\
Stiffness coefficient of the bellows & $k_{\mathrm{b}}$ & 968.8 N/m \\
Damping coefficient of the bellows & $c_{\mathrm{b}}$ & 862.4 N$\cdot$s/m \\
Viscous friction coefficient & $\eta$ & 86.97 N$\cdot$s/m \\
\hline
\multicolumn{3}{c}{\textbf{Fin-Pipe Interaction}} \\
\hline
Spatial pitch of the pipe ridges & $d$ & 0.0173 m \\
Fin deformation stiffness (engaged) & $k_{\mathrm{eng}}$ & 1833.1 N/m \\
Fin deformation stiffness (disengaged) & $k_{\mathrm{dis}}$ & 442.0 N/m \\
Threshold for anchoring position switch & $p_{\mathrm{sw}}$ & 0.0175 m \\
Groove clearance dead-zone width & $\delta_{\text{c}}$ & 0.00753 m \\
\hline
\multicolumn{3}{c}{\textbf{Energetics and actuation model}} \\
\hline
Actuation time constant & $\tau$ & 0.155 s \\
Actuation steady-state gain & $K$ & 0.860 \\
Actuation slack width & $\delta_{\mathrm{s}}$ & 0.008 m\\

Idle baseline power of the robot & $P_{\text{idle}}$ & 0.82 W \\
Lumped actuation power factor & $\alpha_{\mathrm{P}}$ & 3.22 \\
\hline
\end{tabular}
\label{tab1}
\end{center}
\vspace{-1mm}
\end{table}

Table~\ref{tab1} summarizes the measured and identified model parameters used in the simulation and experimental studies presented in this paper. These parameters are obtained from the locomotion, actuation, and energy-model identification procedures reported in this section.

\subsection{Locomotion-Model Identification and Results}
\label{subsec:locomotion_identification}

The hybrid locomotion model is identified using a combination of component-level tests and full locomotion experiments. Parameters associated with body compliance and fin--groove interaction are measured directly, while the dissipation and switching parameters are identified from robot motion in the corrugated pipe. During the locomotion experiments, a camera mounted above the pipe records visual markers attached to the robot, from which the realized robot body length $L(t)$ and the measured motion trajectory are obtained.

The parameter identification is carried out in stages. The lumped masses $M_1$ and $M_2$ are obtained directly by weighing the corresponding robot end modules. The bellows stiffness coefficient $k_{\mathrm{b}}$ is identified from a force--deformation test on the compliant body. The fin--groove interaction function in \eqref{eq:fin_force_piecewise} is identified through a dedicated fin test, yielding the effective clearance width $\delta_{\mathrm{c}}$ and the directional stiffness coefficients $k_{\mathrm{eng}}$ and $k_{\mathrm{dis}}$.

\begin{figure}[t]
    \centering
    \includegraphics[width=1.0\linewidth]{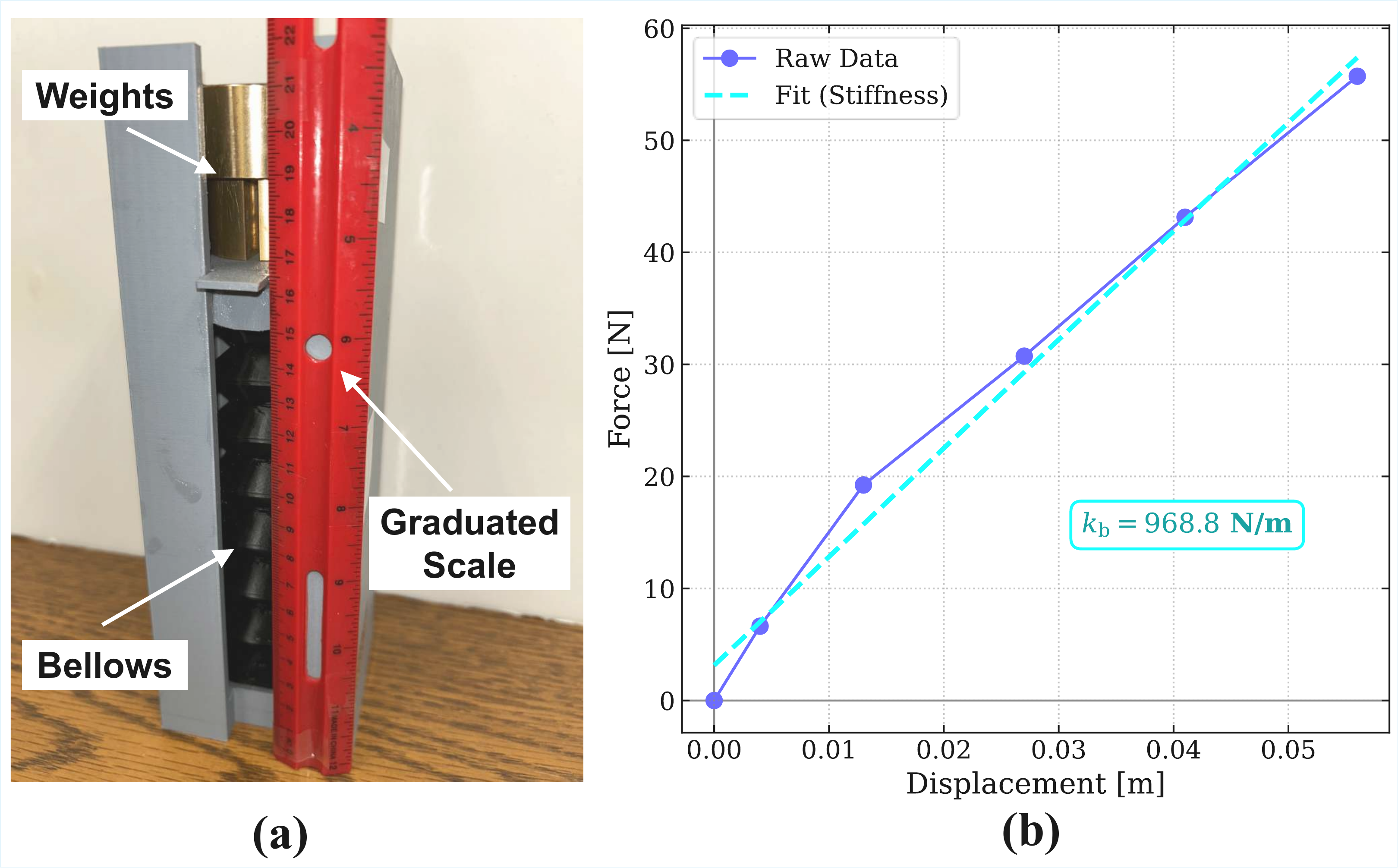}
    \caption{Bellows force--deformation test and linear fit used to identify the stiffness coefficient $k_{\mathrm{b}}$.}
    \label{fig:bellows_stiffness_identification}
    \vspace{-2mm}
\end{figure}

\begin{figure}[t]
    \centering
    \includegraphics[width=1.0\linewidth]{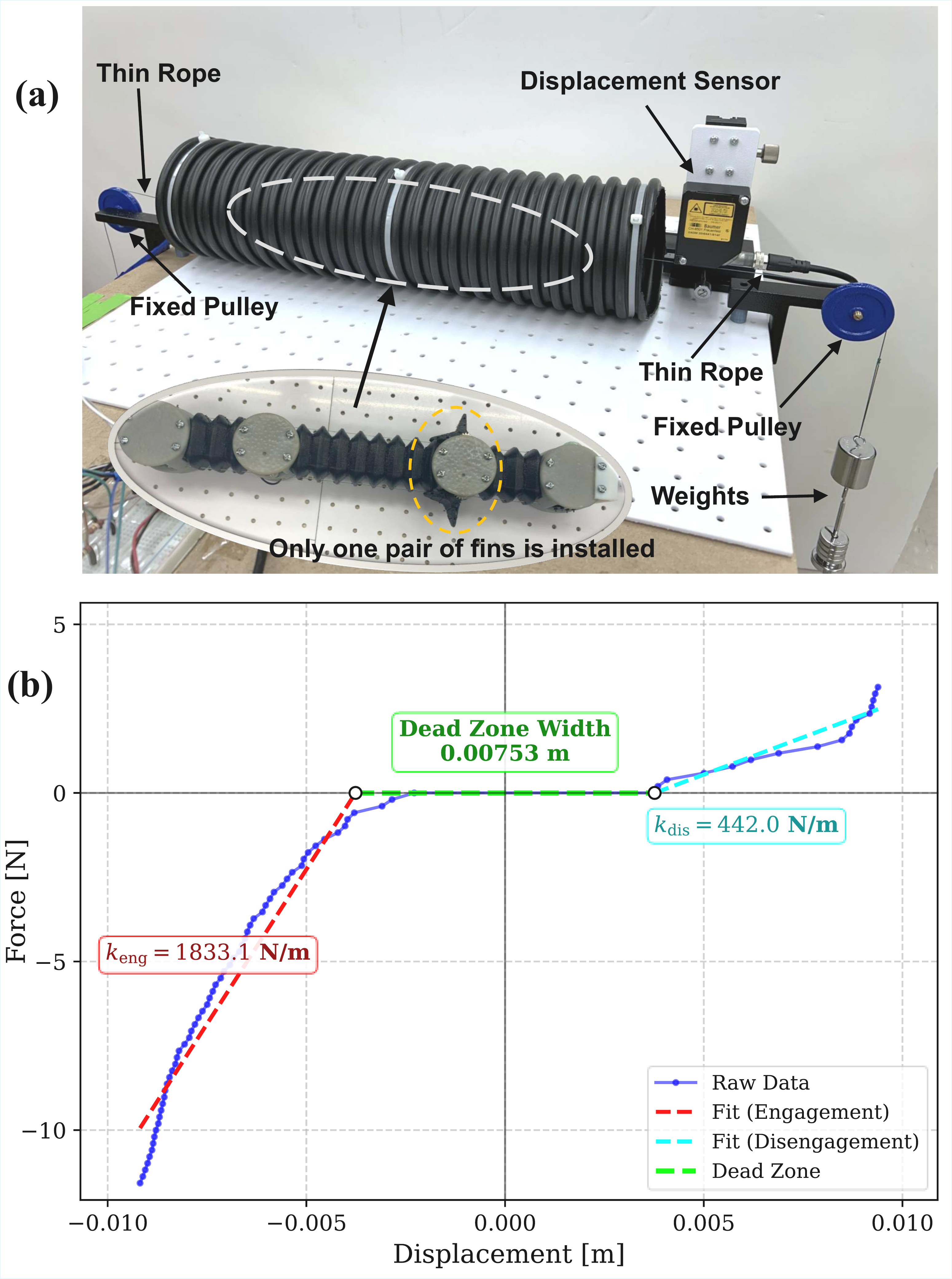}
    \caption{Identification of the clearance-aware fin--groove interaction model. (a) The experimental setup. (b) Experimental results and the identified parameters, $\delta_{\mathrm{c}}$, $k_{\mathrm{eng}}$, and $k_{\mathrm{dis}}$.}
    \label{fig:fin_interaction_identification}
\end{figure}

The remaining parameters, namely the lumped viscous coefficient $\eta$ and the switching threshold $p_{\mathrm{sw}}$, are identified from full locomotion experiments. In this stage, the realized robot body length $L(t)$ is treated as the model input, and the measured locomotion response is represented by the tracked trajectory of the first mass position $x_1(t)$. For each run, the robot traverses the distance of five corrugation ridges, i.e., $5d$.

\begin{figure}[t]
    \centering
    \includegraphics[width=1.0\linewidth]{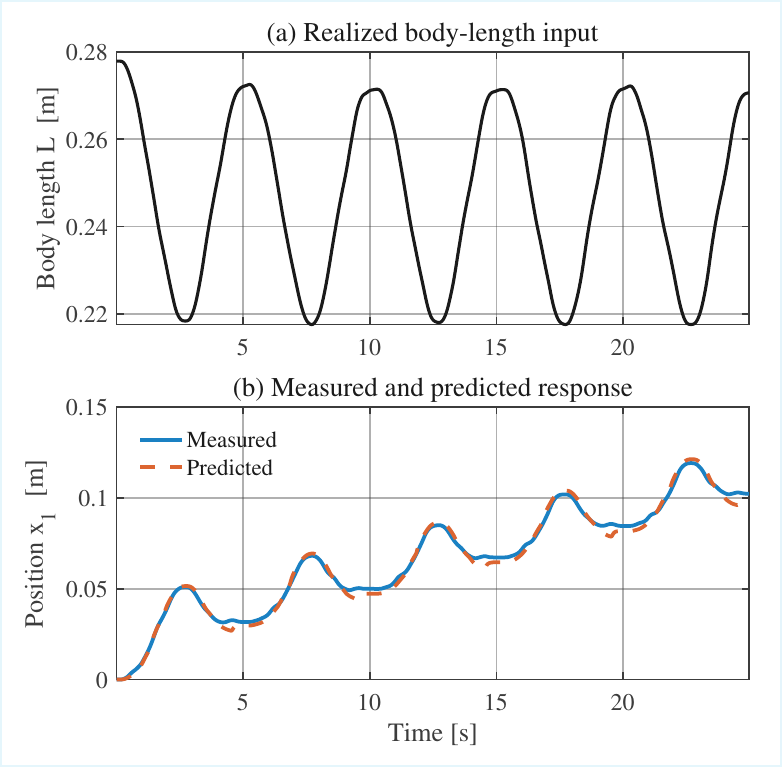}
    \caption{Representative locomotion-model result for the gait with stroke $S=0.07$~m and frequency $f=0.2$~Hz. The figure compares the measured and predicted trajectories of $x_1(t)$ under the experimentally measured realized robot body length $L(t)$.}
    \label{fig:locomotion_identification_validation}
\end{figure}

Let $x_{1,\mathrm{meas}}(t)$ denote the measured trajectory of the first mass block obtained from the motion-tracking data. For a given parameter pair $(\eta,p_{\mathrm{sw}})$, the model-predicted trajectory $x_{1,\mathrm{pred}}(t;\eta,p_{\mathrm{sw}})$ is obtained by integrating the hybrid locomotion model defined by \eqref{eq:continuous_model}--\eqref{eq:A2_switch} under the experimentally measured realized robot body length $L(t)$. The parameters $\eta$ and $p_{\mathrm{sw}}$ are then identified by solving the least-squares problem
\begin{equation}
    \min_{\eta,\;p_{\mathrm{sw}}}
    J_{\mathrm{loc}}(\eta,p_{\mathrm{sw}})
    =
    \frac{1}{N}
    \sum_{k=1}^{N}
    \left(
    x_{1,\mathrm{meas}}(t_k)
    -
    x_{1,\mathrm{pred}}(t_k;\eta,p_{\mathrm{sw}})
    \right)^2,
    \label{eq:locomotion_identification_cost}
\end{equation}
where $N$ is the number of sampled time points over the five-ridge traversal interval. The optimization in \eqref{eq:locomotion_identification_cost} is solved numerically to obtain $(\eta,p_{\mathrm{sw}})$. Figure~\ref{fig:locomotion_identification_validation} shows a representative result for the gait with stroke $S=0.07$~m and frequency $f=0.2$~Hz, where the model is driven by the experimentally measured realized robot body length $L(t)$ and the predicted trajectory is compared with the measured $x_1(t)$.

The close agreement indicates that the identified locomotion model captures the dominant locomotion response of the robot under the tested conditions. This hybrid locomotion model serves as the core motion-prediction component of the overall framework and also yields the cable-force output in \eqref{eq:Fc}, which is used in the energy model. As noted earlier, the bellows damping coefficient $c_{\mathrm{b}}$ is identified later through the energy-model data.

\subsection{Actuation-Model Identification and Results}
\label{subsec:actuation_identification}

The parameters of the slack-aware actuation model are identified using the same experimental campaign used for locomotion-model identification in the preceding subsection, so that the actuation layer is calibrated under the same operating conditions as the locomotion model that it feeds.

Let $\Delta L_{\mathrm{meas}}(t)$ denote the recorded realized body-length change obtained from this experimental campaign, so that $L_{\mathrm{meas}}(t)=L_0+\Delta L_{\mathrm{meas}}(t)$. Here, $\Delta L_{\mathrm{meas}}(t)\le 0$ under the contraction-dominant sign convention adopted in this work. The commanded input $u_{\mathrm{cmd}}(t)$ is generated from the servo command through the nominal rotation-to-length transmission ratio, and the model-predicted response $\Delta L_{\mathrm{pred}}(t;\boldsymbol{\theta}_{\mathrm{act}})$ is obtained by first computing the effective input $u_{\mathrm{eff}}(t)$ via the slack clip in \eqref{eq:slack_clip} and then propagating the first-order dynamics in \eqref{eq:first_order_actuation}.

\begin{figure}[t]
    \centering
    \includegraphics[width=\linewidth]{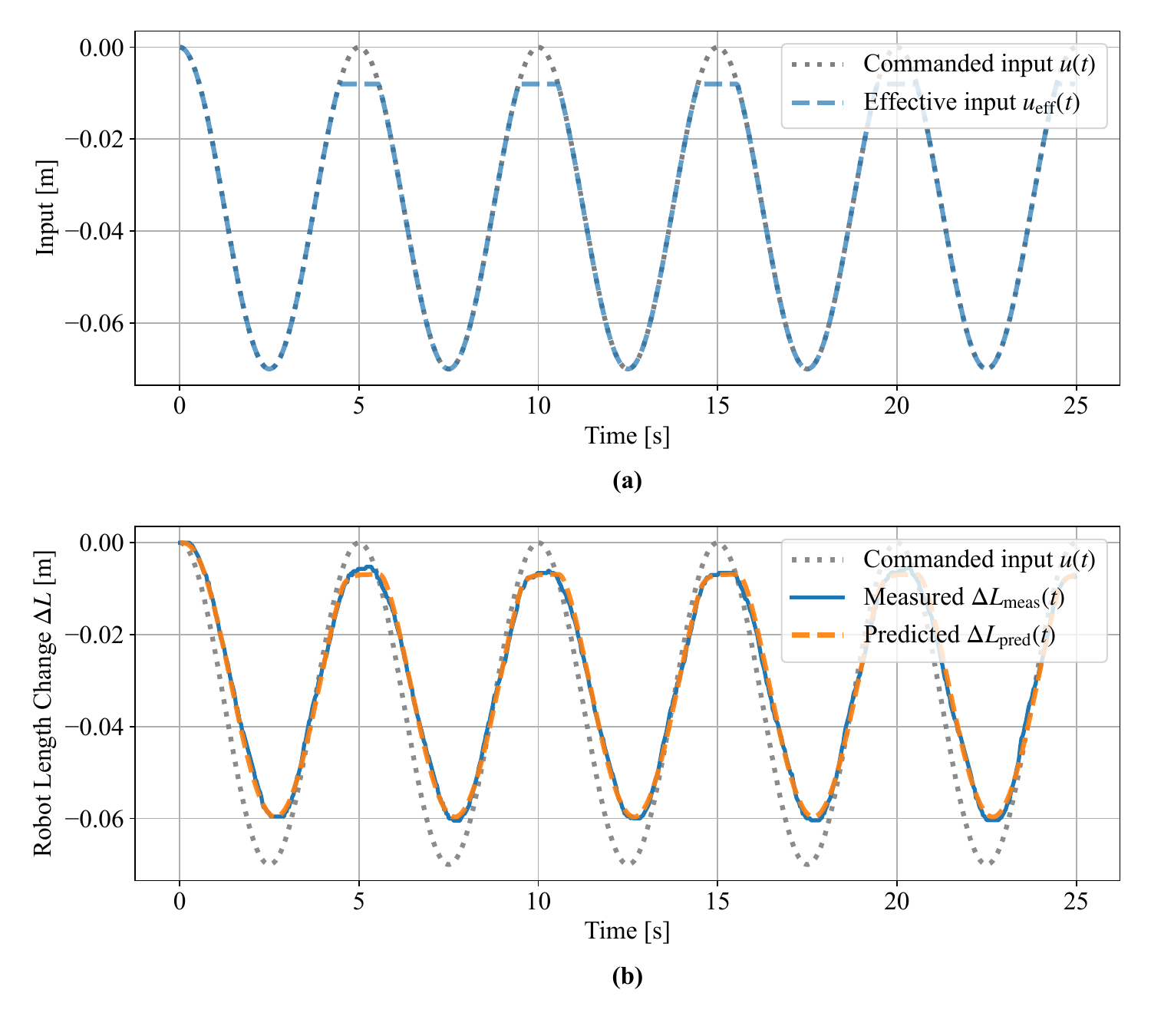}
    \caption{Representative actuation-model result for the gait with stroke $S=0.07$~m and frequency $f=0.2$~Hz. (a) Measured and predicted realized body-length change. (b) Commanded and effective inputs under the identified slack-aware actuation model.}
    \label{fig:actuation_ucmd_ueff}
\end{figure}

The actuation parameter vector is $\boldsymbol{\theta}_{\mathrm{act}}=\left[\delta_{\mathrm{s}},\;K,\;\tau\right]^\top$. These parameters are identified by minimizing the discrepancy between the measured and predicted body-length change over the training dataset:
\begin{equation}
\min_{\boldsymbol{\theta}_{\mathrm{act}}}
J_{\mathrm{act}}(\boldsymbol{\theta}_{\mathrm{act}})
=
\frac{1}{N}
\sum_{k=1}^{N}
\left(
\Delta L_{\mathrm{meas},k}
-
\Delta L_{\mathrm{pred},k}(\boldsymbol{\theta}_{\mathrm{act}})
\right)^2,
\label{eq:actuation_identification_cost}
\end{equation}
where $N$ is the total number of sampled data points used for training. The optimization in \eqref{eq:actuation_identification_cost} is solved numerically using fitting runs from the same experimental campaign. Figure~\ref{fig:actuation_ucmd_ueff} shows a representative result for the gait with stroke $S=0.07$~m and frequency $f=0.2$~Hz. The lower panel compares the commanded input $u_{\mathrm{cmd}}(t)$ and the effective input $u_{\mathrm{eff}}(t)$ inferred by the identified slack-aware actuation model, while the upper panel compares the measured realized body-length change $\Delta L_{\mathrm{meas}}(t)$ and the predicted response $\Delta L_{\mathrm{pred}}(t)$. The close agreement indicates that the proposed actuation model captures both the reversal-dependent transmission effect and the dominant low-order response of the robot body.

These results support the use of the proposed slack-aware actuation model as the front-end of the overall modeling framework. In particular, it provides a physically interpretable mapping from the controller-level commanded input $u_{\mathrm{cmd}}(t)$ to the realized body-length change $\Delta L(t)$ while explicitly capturing the reversal-dependent slack effect induced by cable transmission and pipe loading.

\subsection{Energy-Model Identification and Results}
\label{subsec:energy_identification}

The parameters of the energy model are identified using the same experimental campaign used above for locomotion and actuation modeling. The reported experiments are based on runs in which the robot traverses a fixed distance of five ridges, i.e., $5d$, so that the accumulated energy can be compared consistently across operating conditions.

During each experiment, an INA219 current-and-voltage sensor is used to record the source-side electrical power consumption of the robot system from actuation onset. Let $P_{\mathrm{meas}}(t)$ denote the measured source-side power.

\begin{figure}[t]
    \centering
    \includegraphics[width=1.0\linewidth]{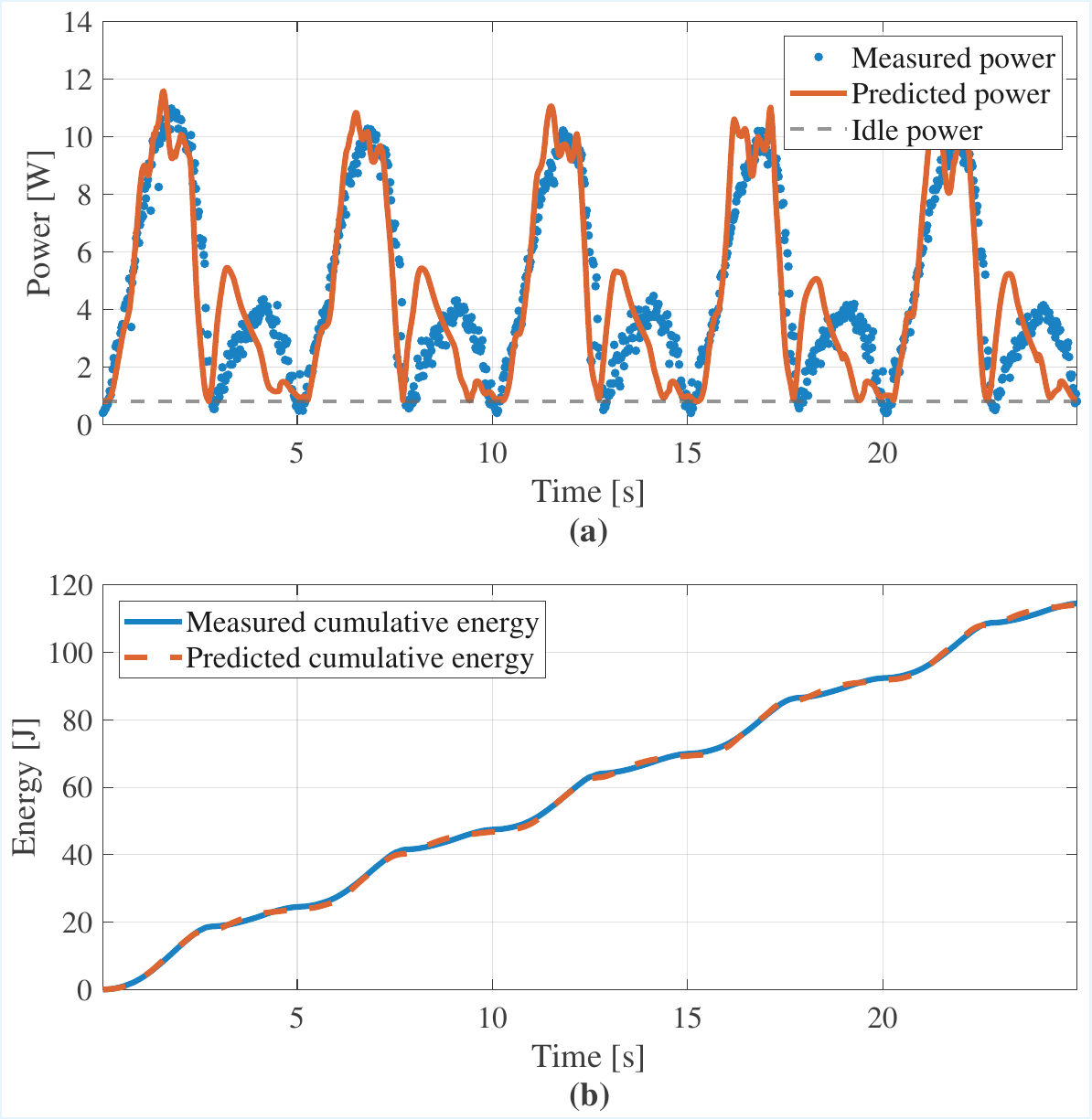}
    \caption{Representative energy-model result for the gait with stroke $S=0.07$~m and frequency $f=0.2$~Hz. (a) Measured and predicted source-side power. (b) Measured and predicted accumulated energy over the five-ridge traversal.}
    \label{fig:energy_validation}
    \vspace{-2mm}
\end{figure}

For each experimental run, the idle power $P_{\mathrm{idle}}$ is estimated directly from the data as the average of the first ten measured power samples:
\begin{equation}
    P_{\mathrm{idle}}
    =
    \frac{1}{10}\sum_{k=1}^{10} P_{\mathrm{meas},k}.
    \label{eq:Pidle_est}
\end{equation}
The measured accumulated energy trajectory $E_{\mathrm{meas}}(t)$ is then obtained by numerical integration of the power signal over the five-ridge traversal interval. For energy-model identification, the realized robot body length $L(t)$ is obtained from physical measurements and numerically differentiated, after smoothing, to obtain $\dot{L}(t)$. Together with the cable force $F_{\mathrm{c}}(t)$ from the locomotion model, these signals are used in \eqref{eq:power_model} to compute the predicted power and the corresponding accumulated energy trajectory. With $P_{\mathrm{idle}}$ fixed from \eqref{eq:Pidle_est}, the remaining energy-model parameters are $c_{\mathrm{b}}$ and $\alpha_{\mathrm{P}}$, which are identified by minimizing the discrepancy between the measured and predicted accumulated energy trajectories over the training dataset:
\begin{equation}
    \min_{\boldsymbol{\theta}_E}
    J_E(\boldsymbol{\theta}_E)
    =
    \frac{1}{M}
    \sum_{i=1}^{M}
    \frac{1}{N_i}
    \sum_{k=1}^{N_i}
    \left(
    E_{\mathrm{meas},i}(t_k)
    -
    E_{\mathrm{pred},i}(t_k;\boldsymbol{\theta}_E)
    \right)^2,
    \label{eq:energy_identification_cost}
\end{equation}
where $M$ is the number of training runs and $N_i$ is the number of sampled time points in the $i$th run. Figure~\ref{fig:energy_validation} shows a representative result for the gait with stroke $S=0.07$~m and frequency $f=0.2$~Hz. The upper panel compares the measured and predicted source-side power, and the lower panel compares the corresponding accumulated energy trajectories over the five-ridge traversal.

These results indicate that the proposed energy model captures the dominant dependence of source-side energy consumption on locomotion actuation over the tested conditions. The identified energy model and its fitted parameters are used in the optimization study and the hardware results of optimized gaits reported next.

\subsection{Price-of-Robustness Analysis for Margin Selection}
\label{subsec:margin_selection_results}

After the locomotion, actuation, and energy models have been identified, the kinematic robustness margin is selected using the price-of-robustness analysis described in Section~IV. The price-of-robustness scan and the subsequent robust optimization are both performed over the admissible gait range $S\in[0.01,\,0.09]$~m and $f\in[0.08,\,0.4]$~Hz. For each candidate value of $\delta_m$, the corresponding optimal cost of transport is computed. Although the final optimization objectives are average speed and average power, the optimal COT provides a convenient scalar indicator for identifying when increasing conservatism begins to incur a substantial performance penalty.

\begin{figure}[t]
    \centering
    \includegraphics[width=\linewidth]{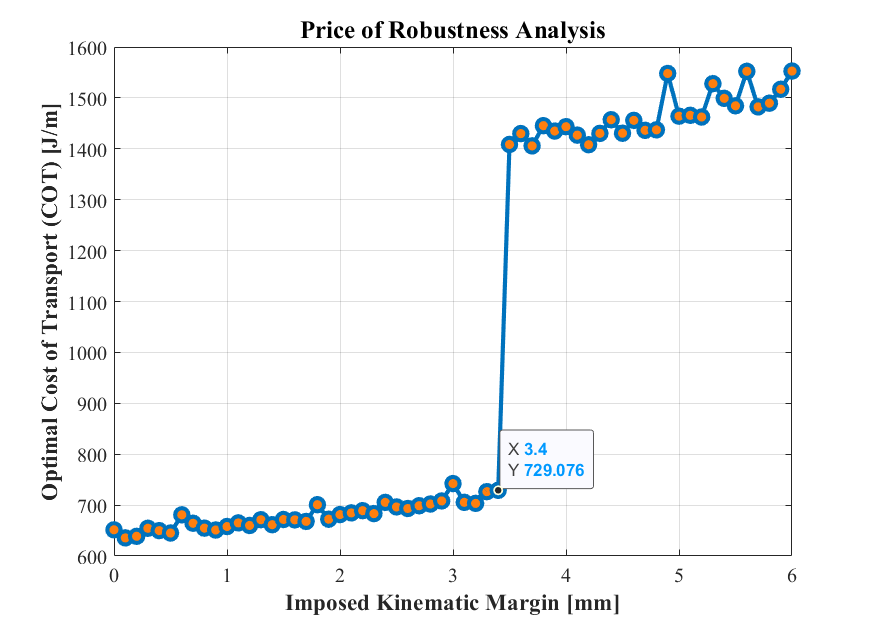}
    \caption{Price-of-robustness analysis used for selecting the kinematic robustness margin. The imposed margin is chosen at the cliff point, where the optimal cost of transport begins to increase abruptly. In the present study, the selected value is $\delta_m=3.4$~mm.}
    \label{fig:robustness_margin_scan}
    \vspace{-1mm}
\end{figure}

Figure~\ref{fig:robustness_margin_scan} shows the resulting price-of-robustness curve. A pronounced cliff point is observed at approximately $\delta_m=3.4$~mm, beyond which the optimal COT increases abruptly. In the present work, this cliff point is therefore selected as the imposed kinematic robustness margin used in the final robust optimization and hardware experiments.

\subsection{Hardware Results of Selected Robustly Optimized Gaits}
\label{subsec:pareto_transferability}

To assess the practical effectiveness of the optimization approach developed in Section~IV, representative solutions are selected from the robust Pareto front and tested on the physical robot. In the present study, the robustness margin is fixed at the selected value $\delta_m=3.4$~mm, and three representative points are chosen to illustrate different operating regimes: the minimum-power point, an intermediate cruising point, and the maximum-speed point. For each selected solution, the corresponding sinusoidal gait in \eqref{eq:problem_formulation_gait} is applied to the robot, and the motion is recorded over five actuation cycles. As in simulation, the first two cycles are treated as transient, and the last three cycles are used to compute the measured average speed and average power.

\begin{figure}[t]
    \centering
    \includegraphics[width=\linewidth]{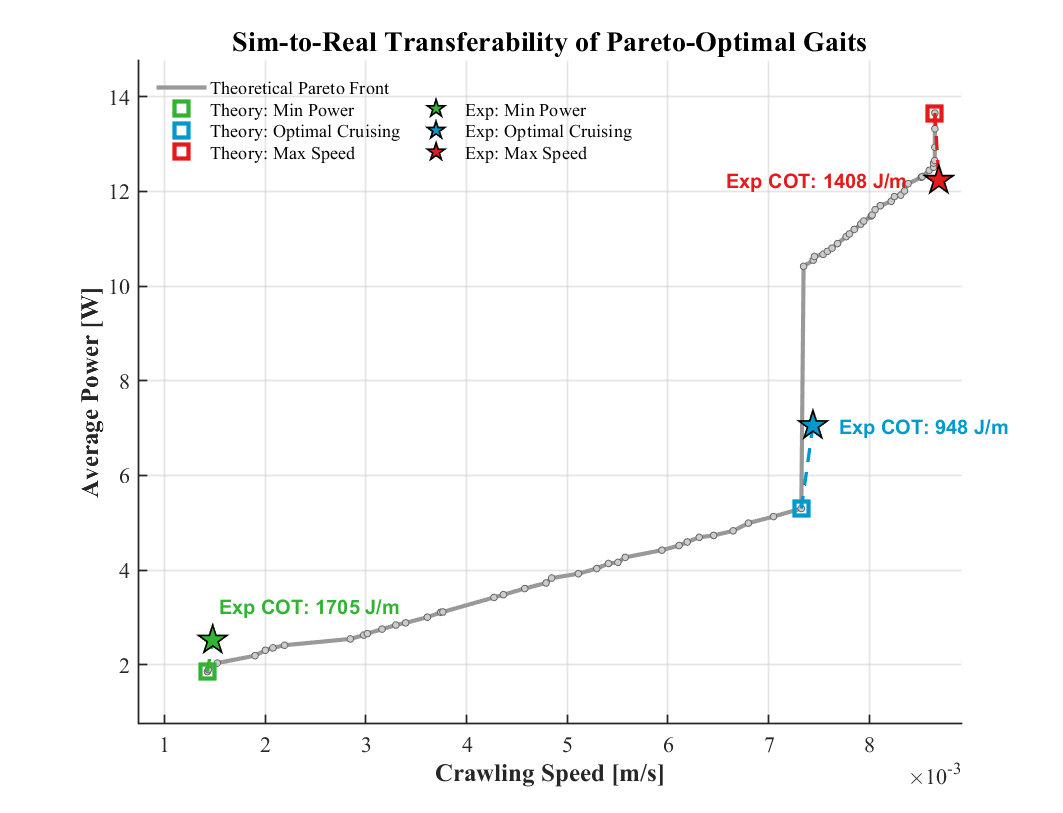}
    \caption{Sim-to-real transferability of the selected robustly optimized gaits. The theoretical Pareto front is shown in the average speed--average power plane, together with the experimental results of the selected minimum-power, cruising, and maximum-speed gaits. The annotated experimental COT values are provided for reference.}
    \label{fig:pareto_transferability}
\end{figure}

Figure~\ref{fig:pareto_transferability} compares the simulated Pareto front with the experimentally measured performance of the selected robust gaits. The theoretical Pareto front is shown in the speed--power plane, and the three corresponding experimental results are superimposed. Although the optimization objectives are average speed and average power, the experimentally measured COT values of the selected points are also annotated in the figure for reference.

The results show meaningful transfer from simulation to hardware under the tested conditions. The experimentally measured points remain qualitatively consistent with the predicted Pareto trade-off, and the minimum-power point, cruising point, and maximum-speed point preserve their intended operating characteristics, indicating that the imposed kinematic robustness margin improves the hardware deployability of the optimized gaits.

These results indicate that the proposed robust gait optimization framework with a kinematic margin improves the hardware deployability of model-based gait design for the worm robot. By explicitly enlarging the switching threshold by $\delta_m$, the formulation reduces the tendency of the optimizer to exploit fragile edge-case solutions near the locomotion-transition boundary.


\section{Conclusion}
\label{sec:conclusion}

This paper presented an experimentally grounded framework for modeling and gait optimization of a compliant worm robot traversing in corrugated pipes. Beyond the dynamic locomotion modeling study, the present work developed a more complete pipeline that connects commanded gait input, realized body-length change, locomotion response, and energetic cost.

First, a hybrid locomotion model was formulated for the robot and experimentally identified using component-level tests and camera-based locomotion experiments. In particular, a clearance-aware fin--groove interaction model was introduced to better capture the directional anchoring behavior of the compliant fins in the corrugated environment. Second, a slack-aware first-order actuation model was developed to describe the mapping from the commanded actuation input to the realized robot body-length change. Third, an energy model was formulated and identified from source-side power measurements, providing an experimentally grounded description of the energetic cost associated with locomotion. Finally, these models were integrated into a robust multi-objective gait optimization framework with a kinematic robustness margin, so that optimized gaits could be selected based on the trade-off between average speed and average power while remaining more deployable in hardware.

The experimental results showed that the proposed framework captures the dominant locomotion and energy behavior of the robot over the tested conditions and supports hardware validation of optimized gaits. In particular, the results highlighted that nominal boundary-seeking optimization can be fragile in physical experiments, and that introducing a kinematic robustness margin provides a practical way to improve the consistency between model-based optimization and hardware execution.

Overall, this work closes the loop from commanded gait to hardware-validated gait optimization for the compliant worm robot by combining experimentally identified locomotion, actuation, and energy models with a margin-based robust optimization framework. The resulting framework provides a basis for systematic gait design of the compliant worm robot platform.

Future work will further improve the framework in several directions. One direction is to refine the locomotion-transition modeling under uncertainty and develop more explicit uncertainty-aware optimization methods. Another is to extend the framework to a broader range of operating conditions, such as curved pipes and pipes with water flow. It is also of interest to investigate whether similar modeling and optimization ideas can be generalized to other compliant robots that rely on anisotropic anchoring and cyclic body deformation for locomotion.

\section*{DECLARATION OF GENERATIVE AI AND AI-ASSISTED TECHNOLOGIES IN THE WRITING PROCESS}
During the preparation of this work, the author(s) used large language model (LLM) tools, specifically ChatGPT, to improve the language and readability of the manuscript. After using this tool/service, the author(s) reviewed and edited the content as needed and take(s) full responsibility for the content of the final publication.

\bibliography{ifacconf}             


\end{document}